\documentclass[letterpaper, 10 pt, conference]{ieeeconf}  

\IEEEoverridecommandlockouts 

\usepackage{times,amsmath,amsfonts,stmaryrd,multirow,subfigure,epsfig,enumerate}
\usepackage{url}
\usepackage{flushend}
\usepackage{cite}
\usepackage{array}
\usepackage{color}
\usepackage{pdfsync}
\usepackage{textcomp}
\usepackage{graphicx}
\usepackage{algorithm}
\usepackage{algcompatible}
\usepackage{textcomp}

\title{\LARGE \bf
Interactive Text2Pickup Network for Natural Language based Human-Robot Collaboration
}
\author{Hyemin Ahn, Sungjoon Choi, Nuri Kim, Geonho Cha, and Songhwai Oh%
\thanks{
H. Ahn, N. Kim, G. Cha, and S. Oh are with the Department of Electrical and Computer Engineering and ASRI,
Seoul National University, Seoul, 08826, Korea
(e-mail: {hyemin.ahn, nuri.kim, geonho.cha}@cpslab.snu.ac.kr, songhwai@snu.ac.kr).
S. Choi is with Kakao Brain, Sungnam, 13494, Korea
(e-mail:sam.choi@kakaobrain.com).
}
}

\begin{document}

\maketitle

\begin{abstract}
In this paper, we propose the Interactive Text2Pickup (IT2P) network
for human-robot collaboration which
enables an effective interaction with a human user despite the
ambiguity in user's commands. 
We focus on the task where a robot is expected to pick up an object
instructed by a human, 
and to interact with the human when the given instruction is vague.
The proposed network understands the command from the human user
and estimates the position of the desired object first.
To handle the inherent ambiguity in human language commands,
a suitable question which can resolve the ambiguity is generated.
The user's answer to the question is combined with the initial command
and given back to the network, resulting in more accurate estimation.
The experiment results show that given unambiguous commands,
the proposed method can estimate the position of the requested object
with an accuracy of 98.49\% based on our test dataset.
Given ambiguous language commands,
we show that the accuracy of the pick up task increases by 1.94 times
after incorporating the information obtained from the interaction.

\end{abstract}

\section{Introduction}
Language is one of the effective means of communication in humans.
However, \textit{``language is the source of misunderstandings,''} as
\textit{Antoine de Saint-Exup\'ery} said since a person may say
ambiguous sentences which can be interpreted in more than one way
\cite{language_ambiguity}. 
In this case, we interact with the speaker
by asking additional questions to reduce the vagueness in the sentence.
This situation can also occur in human-robot collaboration,
where humans can give an ambiguous language command to a robot
and the robot needs to find an appropriate question to understand their intention.
Inspired by this observation, we present a novel method
which enables a robot to handle the ambiguity inherent in the human language command.
Without any preprocessing procedures such as language parsing and object detection,
the proposed method based on neural networks
can deal with the given collaboration task more accurately.

In this paper, we focus on the task where a robot is expected to pick up a specific object
instructed by a human.
There have been several studies related to making robots pick up objects
when a human has given a language command \cite{paul2016, paul2017, whitney2017}.
The methods in \cite{paul2016, paul2017} use a probabilistic graph model
to make a robot understand human language commands and recognize where the object is.
However, these studies do not consider the ambiguity in the human language,
and our goal is to achieve a successful human-robot collaboration
by alleviating the vagueness in the language through the interaction.
\cite{whitney2017} suggests a model
which can make a robot fetch a requested object to a human
and ask if the given language command is ambiguous.
It also focuses on mitigating the ambiguity in the language command by interacting with humans,
but executes an experiment in a simple environment
where three types of six objects are arranged in a single line.

\begin{figure*}[t]
\centering
\includegraphics[width=\linewidth]{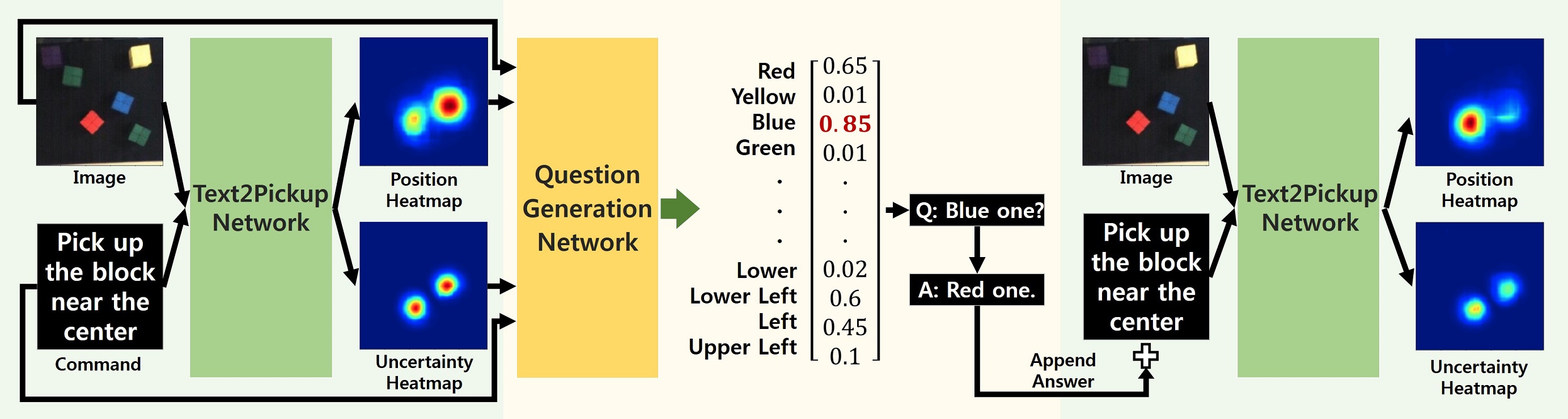}
\caption{
An overview of the proposed Interactive Text2Pickup (IT2P) network.
An image capturing the environment
and a language command from a human user
are provided as the input to the Text2Pickup network.
Based on the input,
a Text2Pickup network generates a position heatmap 
and an uncertainty heatmap.
A question generation network receives the image, the language command,
and two generated heatmaps as the input and generates an appropriate question 
to clarify the human intention.
The human response to the question is appended to the initial language command
and given back to the Text2Pickup network, producing better estimation results.
}
\label{fig:overall}
\end{figure*}

In this paper, 
we propose the Interactive Text2Pickup (IT2P) network,
which consists of a Text2Pickup network and a question generation network.
Once receiving the image of the environment and the language command from a human user,
the Text2Pickup network generates a position heatmap,
which is a two-dimensional distribution with values indicating the
confidence in the position of the desired object,
as well as an uncertainty heatmap,
which is a two-dimensional distribution modeling the uncertainty in the generated position heatmap.
By training the Text2Pickup network in an end-to-end manner,
we remove the need of the preprocessing step, such as language parsing
and object detection. 
The trained Text2Pickup network can handle the task flexibly
when an input command is complicated or multiple objects are placed in
various ways. 

If the given initial language command is ambiguous,
the question generation network decides which question to ask.
We assume that possible questions are predetermined
by the color and the position of the object (e.g., Red one?, Rightmost one?).
However, even with the set of predefined questions,
choosing an appropriate question is challenging
since the question needs to be related to the given situation and not overlap with the information
which has been already provided from the initial language command.
Regarding this, 
the proposed question generation network generates a suitable question,
which can efficiently query more information about the requested object.
Once the human responses to the generated question,
the answer is accumulated to the initial language command
and given back to the Text2Pickup network for producing a better prediction result.

We show that the Text2Pickup network
can better locate the requested object compared to a simple baseline network.
If a given language command is ambiguous, 
the proposed Interactive Text2Pickup (IT2P) network,
which incorporates additional information obtained from the interaction,
outperforms the single Text2Pickup network in terms of the accuracy.
In the real experiment, we have applied the proposed network using a Baxter robot.
For training and test of the network,
we have collected 477 images taken with the camera on the Baxter arm,
and 27,468 language commands for ordering the robot to pick up different objects.

The goal of the proposed method
can be considered the same as the Visual Question Answering (VQA) task \cite{vqa},
whose goal is to construct a model that can generate an appropriate answer to the image-related question.
However, our goal is to build a model
that can generate a question for alleviating the ambiguity in a given language command,
enabling a successful human-robot collaboration.
Note that finding an answer to a given image-related question,
and generating a question to mitigating the ambiguity of
a given image-related language command are different tasks.

The remainder of the paper is structured as follows.
The proposed Interactive Text2Pickup (IT2P) network
is described in Section~\ref{sec:net}.
Section~\ref{sec:struct} describes the Text2Pickup network
and the question generation network.
Section~\ref{sec:exp} shows the results when the proposed IT2P network 
is applied to a test dataset.
We present the quantitative results
showing that the Text2Pickup network infers the object position
better than the baseline single Text2Pickup network.
The demonstration of the proposed method using a Baxter robot is also
provided.

\section{Interactive Text2Pickup Network} \label{sec:net}

Let $\mathbf{I} \in \mathcal{I} \subset \mathbb{R}^{n_i \times n_i \times 3}$ be
an image of the environment shared between a robot and a human user,
where $n_i$ is the size of the image,
and $\mathcal{I}$ is a set of possible images from the environment.
The language command from a human user is represented as
$\mathbf{C}=\{w_1^c, \ldots, w_{n_c}^c\} \in \mathcal{C}$,
where $w_k^c \in \mathbb{R}^d$ denotes the one-hot vector representation of the $k$-th word,
$d$ is the size of vocabulary,
$n_c$ is the number of words in the command,
and $\mathcal{C}$ is the set of possible commands from humans.

As shown in Figure~\ref{fig:overall},
the proposed Interactive Text2Pickup (IT2P) network consists of
a Text2Pickup network and a question generation network.
A Text2Pickup network $\mathcal{F}_T$
takes an image and a language command as the input
and generates two heatmaps as follows: 
\begin{equation}
	\mathcal{F}_T: \mathcal{I} \times \mathcal{C} 
	\rightarrow
	\mathbb{R}^{n_m \times n_m \times 2}, \;
	\mathcal{F}_T(\mathbf{I}, \mathbf{C}) = \{\mathbf{M}_p, \mathbf{M}_u\}
\nonumber
\end{equation}
where $\mathbf{M}_p \in \mathbb{R}^{n_m \times n_m}$ is a position heatmap
which estimates the location of the requested object,
and $\mathbf{M}_u \in \mathbb{R}^{n_m \times n_m}$ is an uncertainty heatmap
which models the uncertainty in $\mathbf{M}_p$.

In this paper, we assume that types of questions that a robot can ask
are predetermined,
such that questions are related to the color or the position of an object,
or inquire whether the current estimation is correct or not.
Let $\mathcal{Q}=\{q_1, \ldots q_{n_q}\}$ be a set of $n_q$ predefined questions.
The question generation network $\mathcal{F}_Q$ determines which question to ask.
As depicted in Figure~\ref{fig:overall},
based on the image $\mathbf{I}$, the language command $\mathbf{C}$,
and generated heatmaps, $\mathbf{M}_p$ and $\mathbf{M}_u$,
$\mathcal{F}_Q$ generates a weight vector
$\mathbf{V}_Q =[v_1, \ldots, v_{n_q}]^T \in \mathbb{R}^{n_q}$,
which can be denoted as follows:
\begin{align}
	\mathcal{F}_Q:  \mathcal{I} & \times \mathcal{C} \times \mathbb{R}^{n_m \times n_m \times 2}
	\rightarrow \mathbb{R}^{n_q}
\nonumber \\
	\mathcal{F}_Q & \big(\mathbf{I}, \mathbf{C}, \{\mathbf{M}_p, \mathbf{M}_u\} \big) = \mathbf{V}_Q
\nonumber
\end{align}

The predefined questions are sorted based on the weight values in $\mathbf{V}_Q$,
and the question with the highest weight value is chosen and asked to the human.
Let $\mathbf{A} = \{w_1^a, \ldots, w_{n_a}^a \}$ be an answer from the human user to the question,
where $w_k^a$ is the $k$-th word consisting the answer and
$n_a$ is the number of words in the answer.
The answer is appended to the initial language command $\mathbf{C}$
and the augmented command $\mathbf{C}^+ = \{w_1^c, \ldots, w_{n_c}^c, w_1^a, \ldots, w_{n_a}^a \}$ is generated.
Based on this,  $\mathcal{F}_T$
generates a better estimation result $\mathcal{F}_T(\mathbf{I}, \mathbf{C}^+)$
as shown in Figure~\ref{fig:overall}.

\section{Network Structure} \label{sec:struct}
\subsection{Text2Pickup Network} \label{sec:hgn}

\begin{figure}[t]
\centering
\includegraphics[width=\linewidth]{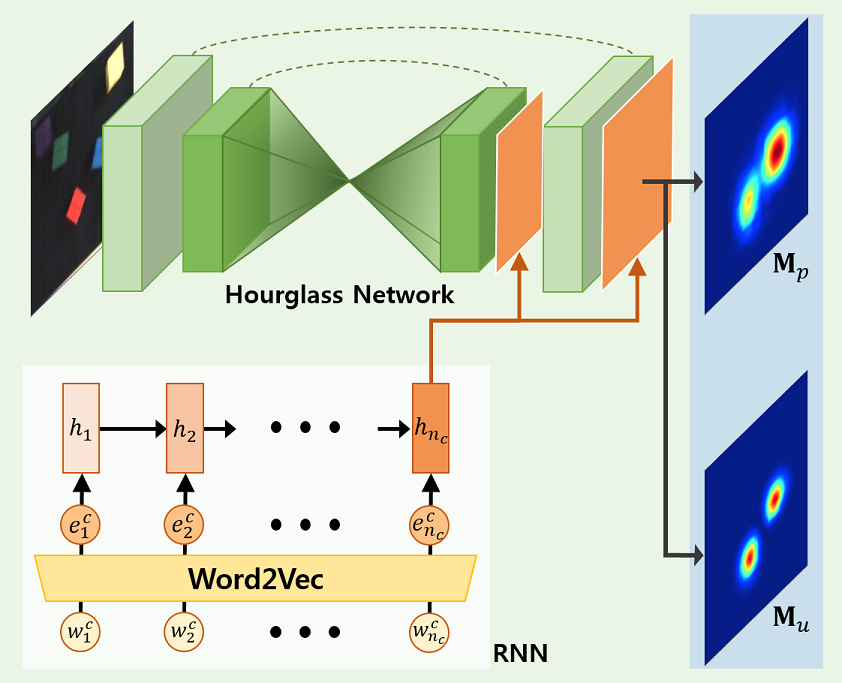}
\caption{
The structure of the Text2Pickup network,
which is composed of a single Hourglass network \cite{hourglass}
combined with a RNN \cite{lstm}.
When the Hourglass network upsamples a set of encoded image features,
the last hidden state vector $h_{n_c}$ of the RNN
is delivered prior to each upsampling procedure.
After that, the position heatmap $\mathbf{M}_p$ and the uncertainty heatmap $\mathbf{M}_u$
are generated
(see equations (\ref{eqn:heatmap_1})-(\ref{eqn:elementwise})).
}
\label{fig:hgn}
\end{figure}
A Text2Pickup network $\mathcal{F}_T$
takes an image $\mathbf{I}$ and a language command $\mathbf{C}$
and generates a position heatmap $\mathbf{M}_p$
and an uncertainty heatmap $\mathbf{M}_u$.
We carefully model the architecture of $\mathcal{F}_T$ 
to model the relationship between $\mathbf{I}$ and $\mathbf{C}$ better.
Specifically, the proposed $\mathcal{F}_T$
adapts an Hourglass network \cite{hourglass},
which is mixed with a recurrent neural network (RNN) \cite{lstm} as shown in Figure~\ref{fig:hgn}.

The Hourglass network processes the input image
$\mathbf{I} \in \mathbb{R}^{n_i \times n_i \times 3}$ first,
and the size of processed image $\mathbf{I}_p \in \mathbb{R}^{n_m \times n_m \times 3}$
becomes the same as the heatmap.
On $\mathbf{I}_p$, 
the Hourglass network performs the process of max-pooling after the residual module \cite{residual} repeatedly 
to encode a set of image features from the large resolution to the low resolution.
After repeating this bottom-up process for several times, 
the top-down process which upsamples the encoded image features is executed
as many times as the bottom-up process has been executed,
so that the size of the generated heatmap can be equal to $\mathbf{I}_p$.
The generated heatmap is used after being resized to the initial input image size $n_i$.
After every upsampling procedure,
the image features encoded from the bottom-up process
are passed to the result of the top-down process result by a skip connection,
which is indicated by green dotted lines in Figure~\ref{fig:hgn}.
Interested readers are encouraged to read \cite{hourglass}.

However, since the Hourglass network can only process images,
a human language command which contains the location information of an object cannot be incorporated.
Therefore, as shown in Figure~\ref{fig:hgn}, 
we combine the Hourglass network with a RNN,
which can learn features of sequential data $\mathbf{C}=\{w_1^c, \ldots, w_{n_c}^c\}$.
Before encoding a feature from $\mathbf{C}$, 
we encode all words in $\mathbf{C}$
into a set of word embedding vectors $\{e_1^c, \ldots, e_{n_c}^c \}$,
based on the word2vec model \cite{word2vec}.
Here, $e_k^c \in \mathbb{R}^{n_e}$ is the word embedding representation of $w_k^c \in \mathbb{R}^{d}$,
such that $e_k^c = E w_k^c$, where $E \in \mathbb{R}^{n_e \times d}$ is a word embedding matrix.
Regarding this, we use a pretrained word2vec model provided in \cite{word2vec_model},
whose vector dimension is $n_e = 300$.

The word embedding representation of $\mathbf{C}$ is encoded
into the hidden states of a long short-term memory (LSTM) cell of the RNN \cite{lstm}. 
Let $\mathbf{h} =\{h_1, \ldots, h_{n_c}\}$
be a set of hidden state vectors of the RNN, where
\begin{equation} \label{eqn:lstm}
h_t = \gamma_t [ h_{t-1}, e_t ] 
\end{equation}
Here, $\gamma_t$ is a nonlinear function operating in a LSTM cell.
For more details of how $\gamma_t$ operates, 
we encourage readers to refer the original paper \cite{lstm}.
As shown in Figure~\ref{fig:hgn},
the last hidden state $h_{n_c}$ is delivered to the Hourglass network.
The delivered $h_{n_c}$ is reshaped and concatenated
to the image features before the Hourglass network executes each upsampling process.

Based on the following loss function,
$\mathcal{F}_T$ is trained to generate the prediction of a position heatmap $\hat{\mathbf{M}}_e$:
\begin{equation} \label{eqn:loss_hgn}
\mathcal{L}_{T2P} = \| \mathbf{M}_p^{gt} - \hat{\mathbf{M}}_e \|_2^2,
\end{equation}
where $\mathbf{M}_p^{gt}$ denotes the ground truth position heatmap in the training dataset.

From the trained $\mathcal{F}_T$,
$\mathbf{M}_p$ and $\mathbf{M}_u$ are generated based on the method presented in \cite{uncertainty}.
To be specific,
a set of $T$ predictions of the position heatmap $\{ \hat{\mathbf{M}}_p^1 \ldots \hat{\mathbf{M}}_p^T \}$
is sampled from the trained $\mathcal{F}_T$
with a dropout applied at a rate of $0<d_r<1$.
Based on this,
the position heatmap $\mathbf{M}_p$ and the
uncertainty heatmap $\mathbf{M}_u$ are obtained as follows.
\begin{align}
	\mathbf{M}_p &=  \frac{1}{T} \sum_{t=1}^{T} \hat{\mathbf{M}}_p^t,
\label{eqn:heatmap_1}
\\
	{Var}_p &= \frac{1}{T}  \sum_{t=1}^{T} (\hat{\mathbf{M}}_p^t \circ \hat{\mathbf{M}}_p^t ) -
					      				 \mathbf{M}_p \circ \mathbf{M}_p, 
\\
	\mathbf{M}_u^{ij} &= \sqrt{{Var}_p^{ij}}, \,\, 1 \leq i, j \leq n_m.
\label{eqn:elementwise}
\end{align}   
Here, $\circ$ represents the element-wise production of matrices,
and ${Var}_p$ denotes the predictive variance of the model
based on the method presented in \cite{uncertainty}.
Each element in $\mathbf{M}_u$ is the square root of each element in ${Var}_p$,
and represents the standard deviation of each element in $\mathbf{M}_p$.

\subsection{Question Generation Network} \label{sec:qgn}

We assume that there are $n_q$ questions $Q = \{q_1, \ldots, q_{n_q}\}$ that can be asked to humans.
To choose which question to ask,
a question generation network $\mathcal{F}_Q$ first multiplies $\mathbf{M}_u$ by a coefficient $\beta$
and adds it to $\mathbf{M}_p$ as follows:
\begin{equation} \label{eqn:cmap}
\mathbf{M}_c = \mathbf{M}_p + \beta \mathbf{M}_u
\end{equation}
Note that each element in $\mathbf{M}_u$
represents the standard deviation of each element in $\mathbf{M}_p$.
Building $\mathbf{M}_c$ by the equation (\ref{eqn:cmap}) is inspired by
the Gaussian process upper confidence bound (GP-UCB) \cite{gp-ucb} method,
which solves an exploration-exploitation tradeoff by considering both the estimation mean and its uncertainty.
Therefore, we claim that each element in $\mathbf{M}_c$
represents the upper confidence bound of each element in $\mathbf{M}_p$.

\begin{figure}[t]
\centering
\includegraphics[width=\linewidth]{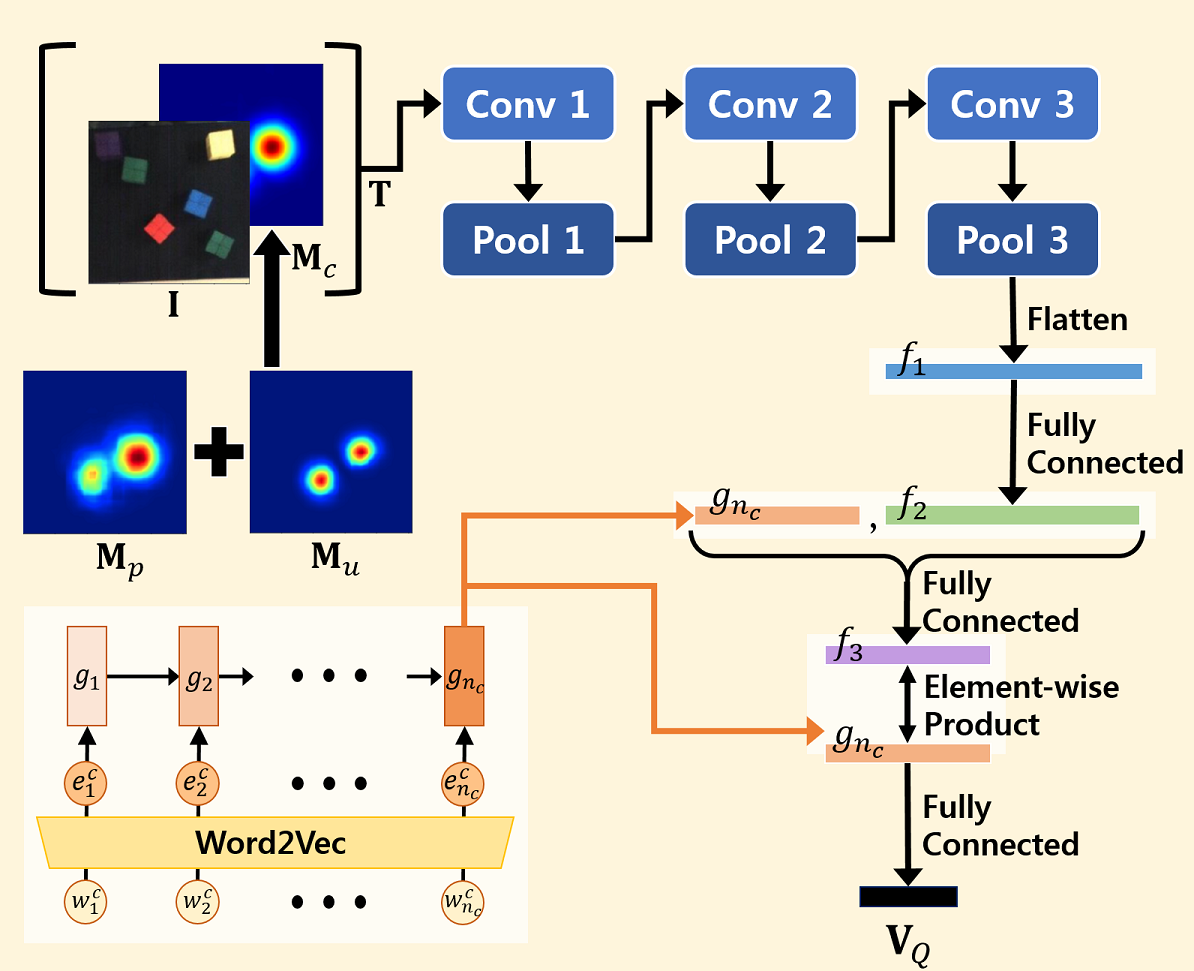}
\caption{
The structure of the question generation network which is a CNN \cite{cnn}
combined with a RNN \cite{lstm}.
The structure of the network is designed to efficiently model the relationship between the image and the language for estimating the value of $\mathbf{V}_Q$ better.
}
\label{fig:qgn}
\end{figure}

The input image $\mathbf{I} \in \mathbb{R}^{n_i \times n_i \times 3}$ is resized
to the size of $\mathbf{M}_c \in \mathbb{R}^{n_m \times n_m}$,
and concatenated to $\mathbf{M}_c$ to generate the tensor
$\mathbf{T} \in  \mathbb{R}^{n_m \times n_m \times 4}$
as shown in Figure~\ref{fig:qgn}.
A question generation network $\mathcal{F}_Q$ is a convolutional neural network (CNN) \cite{cnn}
combined with a RNN.
As shown in Figure~\ref{fig:qgn}, the CNN module
composed of three convolutional and max-pooling layers
encodes the information in $\mathbf{T}$ to a feature vector $f_1$
(the blue rectangular box in Figure~\ref{fig:qgn}).
The RNN module in $\mathcal{F}_Q$ encodes the information in $\mathbf{C}$
as $\mathcal{F}_T$ does (see equation (\ref{eqn:lstm})),
in order to prevent asking the information that has been provided from $\mathbf{C}$.
The last hidden state of the RNN module $g_{n_c}$ (the orange rectangular in Figure~\ref{fig:hgn})
is used as a language feature.

To generate the output vector $\mathbf{V}_Q$ which determines a question to ask,
$\mathcal{F}_Q$ properly combines the image and language features.
To be specific, 
$f_2$ (the green rectangular box in Figure~\ref{fig:hgn}) is generated
by passing $f_1$ to the fully connected layer,
and it is concatenated with $g_{n_c}$.
The concatenated vector passes through a fully connected layer, 
and a feature vector $f_3$ (the purple rectangular in Figure~\ref{fig:hgn})
which has the same size as $g_{n_c}$ is generated.
For more efficient combination of the information in $\mathbf{I}$ and $\mathbf{C}$,
an element-wise product between the $f_3$ and $g_{n_c}$ is performed.
The result is passed to a fully connected layer,
and the output weight vector $\mathbf{V}_Q = [v_1, \ldots, v_{n_q}]^T \in \mathbb{R}^{n_q}$ is generated.
This element-wise product based method is inspired by \cite{cgans},
and it allows $\mathcal{F}_Q$ to better model the relationship between the image and the language,
resulting the better estimation of $\mathbf{V}_Q$.

\subsection{Implementation Details}

The size of the input image $\mathbf{I} \in \mathbb{R}^{n_i \times n_i \times 3}$ is $256$,
and the size of the generated heatmaps $\mathbf{M}_p, \mathbf{M}_u \in \mathbb{R}^{n_m \times n_m}$
is $64$.
For an Hourglass network in the Text2Pickup network $\mathcal{F}_T$,
the bottom-up and top-down process are repeated for four times
and 256 features are used in the residual module.
The top-down process which upsamples the image features encoded from the bottom-up process
is also performed four times.
For the RNN in $\mathcal{F}_T$, the dimension of the hidden state vector $h_t$ is 256.
In order to train $\mathcal{F}_T$, we set the number of epochs 300 with the batch size of eight.
The Adam optimizer \cite{adam} is used to minimize the loss function $\mathcal{L}_{T2P}$,
and a learning rate is set to $10^{-5}$.
When yielding $\mathbf{M}_p$ and $\mathbf{M}_u$
(see equations (\ref{eqn:heatmap_1})-(\ref{eqn:elementwise})),
a dropout rate is set to $d_r=0.1$ and the number of samples is set to $T=100$.

For a question generation network $\mathcal{F}_Q$,
the coefficient value in (\ref{eqn:cmap}) is $\beta=2$.
For its CNN module, the first, second, and third convolutional layers have 16, 32, and 64 filters,
and the size of all filters is 3 by 3.
The max-pooling layer uses a filter size of 2 and a stride of 2.
For the RNN of $\mathcal{F}_Q$, the dimension of the hidden state vector $g_t$ is 256.
In order to train $\mathcal{F}_Q$, we set the number of epochs as $1,000$ with the batch size of eight.
For the loss function, the sparse softmax cross entropy loss function provided by \cite{qgn_loss}
has been used with the learning rate of $10^{-5}$.
All values of these parameters are also chosen empirically\footnote{https://github.com/hiddenmaze/InteractivePickup}.

\section{Experiment} \label{sec:exp}
\subsection{Dataset} \label{sec:data}

For training and test of the Interactive Text2Pickup (IT2P) network,
we have collected a dataset of images capturing the environment observed by a robot.
Each image contains three to six blocks of five colors, 
with up to two blocks of the same color.
Among $477$ images of blocks placed in various ways,
$455$ images are used as a training dataset
and $22$ images are used as a test dataset.
The images were obtained from the camera on the arm of a Baxter robot
to make the proposed network work well in the real environment.

\begin{figure}[t]
\centering
\includegraphics[width=\linewidth]{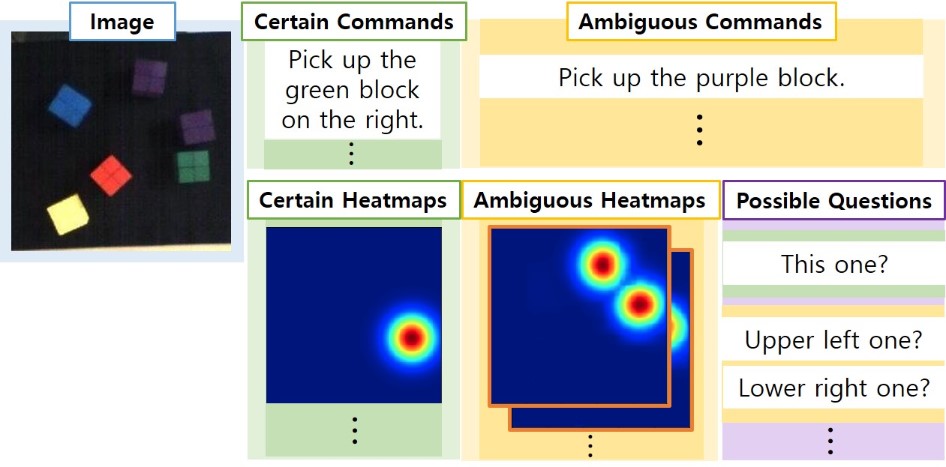}
\caption{
The dataset for training the proposed Interactive Text2Pickup (IT2P) network.
For training the Text2Pickup network, 
an image, an unambiguous language command,
and an unambiguous heatmap indicating the target block position are used as data.
For the input training data of the question generation network,
we have additionally collected ambiguous language commands
and generated heatmaps  $\mathbf{M}_p$, $\mathbf{M}_u$
from the trained Text2Pickup network. 
For the output training data of the question generation network,
we have collected possible questions that a robot can inquire.
}
\label{fig:dataset}
\end{figure}

For the Text2Pickup network,
we collected $20,349$ unambiguous language commands
which clearly specify the desired block.
These language commands are composed of
combinations of representations related to
position (e.g., rightmost, upper, middle),
color (e.g., red, blue, yellow),
and relative position (e.g., between two purple blocks) of the block.
As shown in Figure~\ref{fig:dataset},
each unambiguous language command is paired with an unambiguous heatmap indicating the position of the target block.
Each unambiguous heatmap is generated based on the two-dimensional multivariate Gaussian distribution,
whose mean value is at the center of the target block
with variance of one.
With images, unambiguous language commands, and corresponding unambiguous heatmaps,
the Text2Pickup network can be trained.

For the question generation network, 
the input dataset needs to contain images, language commands,
and heatmaps $\mathbf{M}_p$, $\mathbf{M}_u$ generated from the trained Text2Pickup network.
Regarding this,
we collected $7,119$ ambiguous language commands and
sampled $1,394$ unambiguous language commands from the dataset for the Text2Pickup network.
With collected language commands, $\mathbf{M}_p$, $\mathbf{M}_u$ are generated
from the trained Text2Pickup network and used as a dataset for the question generation network.

For the output dataset of the question generation network, 
we have collected possible questions that a robot can inquire.
The number of predefined questions is set to 15,
which is composed of five questions about the color of the block (e.g., Blue one? Yellow one?),
and nine questions about the position of the block (e.g., Lower one? Upper one?),
and one question to confirm whether the predicted block is correct or not (e.g., This one?).
Each question is encoded as a one-hot vector
and used as the ground truth value of the output weight vector $\mathbf{V}_Q$.

\subsection{Results from the Text2Pickup Network} \label{sec:exp_heatmap}
\begin{figure*}[t]
\centering
\includegraphics[width=\linewidth]{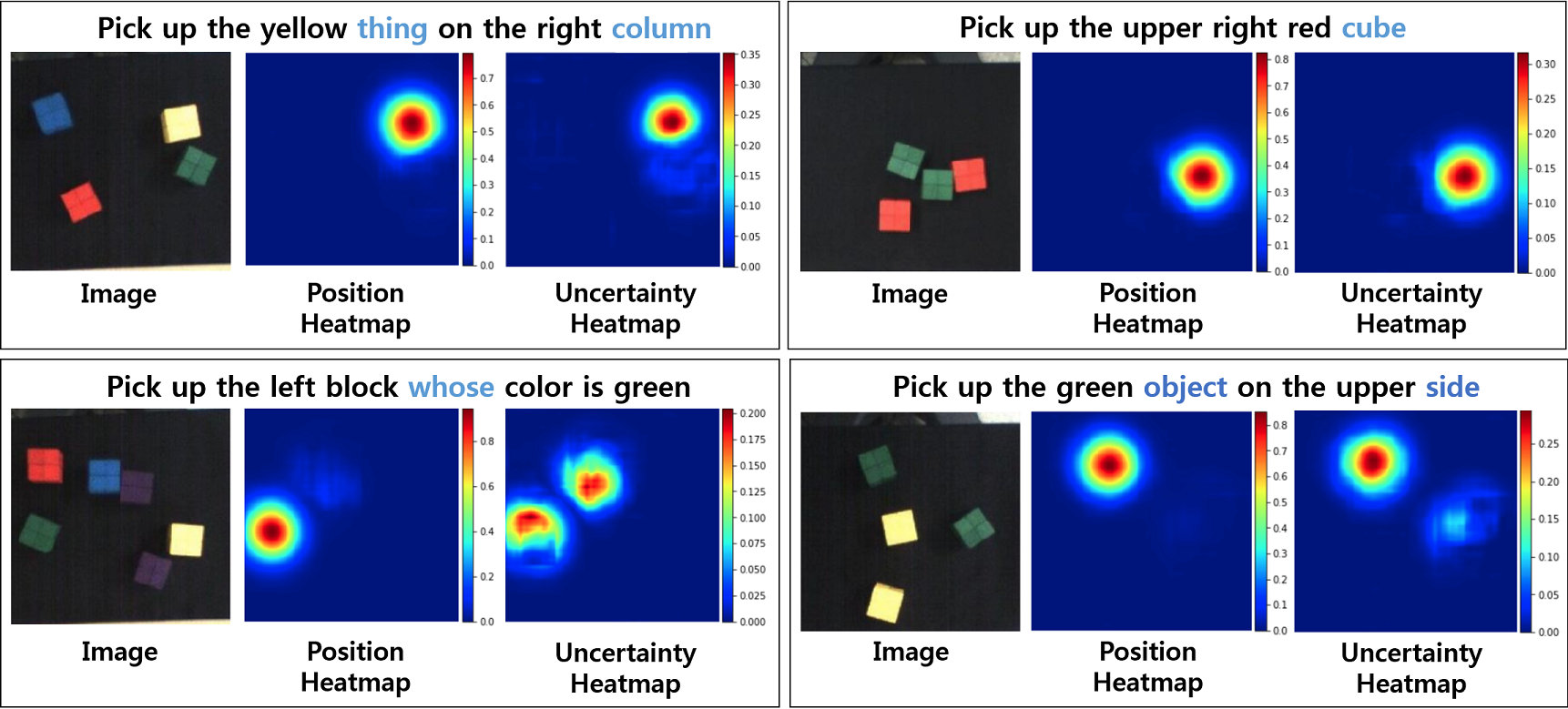}
\caption{
Results from the Text2Pickup network
when four pairs of images and the unambiguous language commands from the test dataset are given.
The results in each rectangle are obtained when the language command (the sentence at the top center of each rectangle)
and the image are given to the trained Text2Pickup network.
In the language command, the word written in blue is a new word which is not included in the training dataset.
}
\label{fig:hgn_result}
\end{figure*}

We first examine how the position and uncertainty heatmaps are generated from the trained Text2Pickup network.
Regarding this, four pairs of images and unambiguous language commands from the test dataset described in Section~\ref{sec:data} are given to the network, and Figure~\ref{fig:hgn_result} shows the results.
In this figure, 
the sentence at the top center of each rectangle represents the given language command,
where the word written in blue is a new word which is not included in the training dataset.
Even with language commands including unseen words,
it is shown that the generated position heatmaps accurately predict the required block position.
This is because the usage of the pretrained word2vec model from \cite{word2vec_model}
helps the network to understand various input words.

The generated uncertainty heatmap has a similar shape
to the position heatmap, showing that the uncertainty heatmap
has a meaningful correlation with the position heatmap
when the language command is unambiguous.
However, 
when the language command is `Pick up the left block whose color is green',
it is shown that the uncertainty value is also high around the position of the blue block.
It is because of the confusion
in the trained Text2Pickup network
when distinguishing between green and blue,
due to the similarity in the RGB space.
In addition,
when the language command is `Pick up the green object on the upper side',
it is shown that the uncertainty value is also high around the position of another green block
since the trained network considers both green blocks.

\subsection{Results from Interaction Scenarios} \label{sec:exp_question}
\begin{figure*}[t]
\centering
\includegraphics[width=\linewidth]{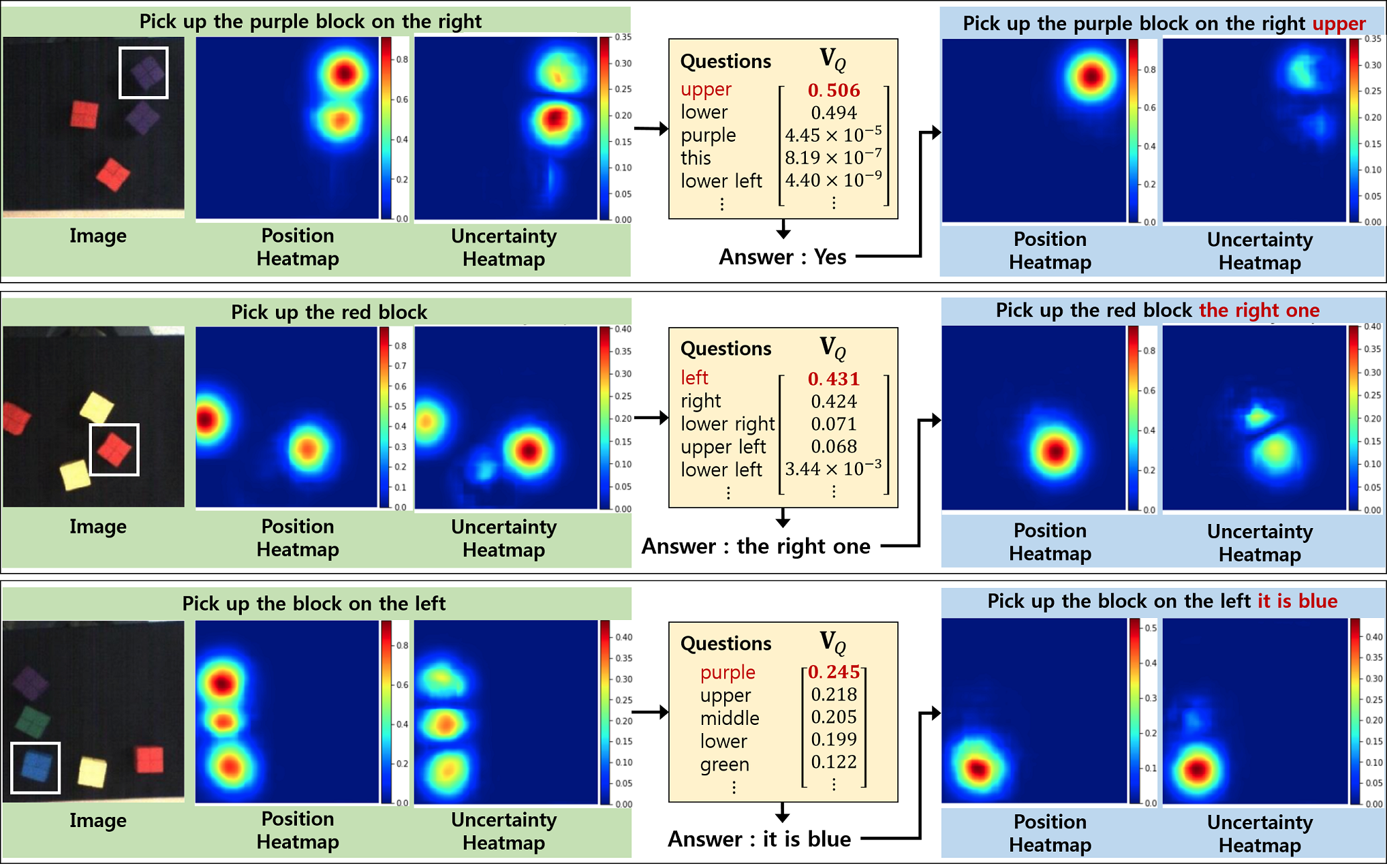}
\caption{
Results from the Interactive Text2Pickup (IT2P) network from interaction scenarios
where the vague language commands are given.
The heatmaps in the green rectangle are obtained results
when images and ambiguous language commands are given as inputs
to the Text2Pickup network.
The white square in the image indicates the block
that the human user actually wanted to ask.
The list in the yellow rectangle
shows five possible questions with high weight values of $\mathbf{V}_Q$,
which are generated from the question generation network.
The question whose weight value is the highest is selected and asked to the human user.
Results in the blue rectangle are generated heatmaps
when a human answer is accumulated to the initial language command
and given back to the Text2Pickup network.
It is shown that the estimation results are improved in terms of accuracy
after obtaining the further information from humans.
}
\label{fig:qgn_result}
\end{figure*}

In this section, we represent how the proposed Interactive Text2Pickup (IT2P) network
works in interaction scenarios, where vague language commands are given.
Figure~\ref{fig:qgn_result} shows the results from three interaction scenarios
which were not included when training the IT2P network.

The heatmaps in the green rectangle are obtained
when images and ambiguous language commands are given
to the Text2Pickup network.
The white square in the image indicates the block
that the human actually wanted to ask.
Based on these two heatmaps, an image, and a language command, 
the question generation network generates a weight vector $\mathbf{V}_Q$
which determines the questions to ask.
In Figure~\ref{fig:qgn_result},
the list in yellow rectangle shows five possible questions with high values of $\mathbf{V}_Q$.
It shows that questions with high weight values
do not inquire the information provided in the initial language command,
and are capable of alleviating the uncertainty in the given language command.

The question with the highest weight value $\mathbf{V}_Q$
is selected and asked to the human user.
The human answer is accumulated to the initial language command and
given back to the Text2Pickup network.
If the answer is `Yes', the question is appended to the initial language command instead of the answer.
Results in the blue rectangle in Figure~\ref{fig:qgn_result}
shows the generated heatmaps after receiving the human answer.
It shows that position heatmaps after the interaction better estimate
the location of the block that the human user originally wanted.
In addition, the value of the uncertainty heatmap is lower overall,
indicating that the resulting position heatmap is more reliable than before.

\subsection{Comparison between the Text2Pickup Network and a Baseline Network} \label{sec:exp_compare_1}
\begin{figure}[t]
\subfigure[]
{
    \centering
    \includegraphics[width=0.575\linewidth]{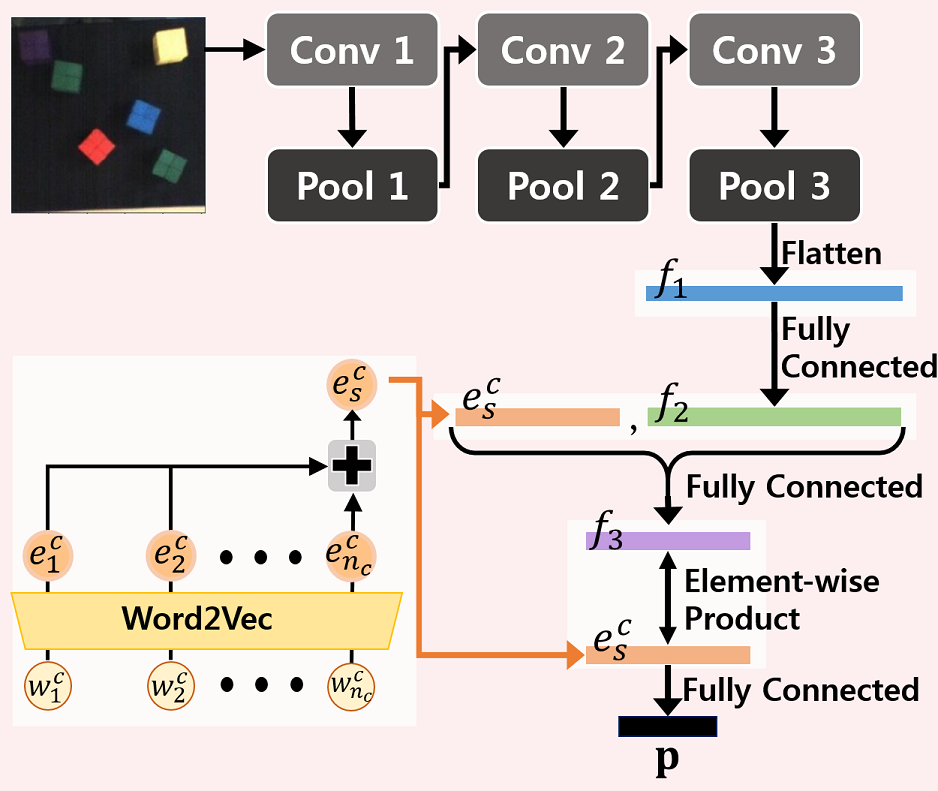}
    \label{fig:baseline}
}%
\subfigure[]
{
    \centering
    \includegraphics[width=0.375\linewidth]{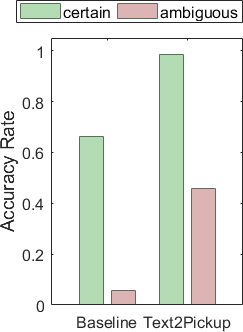}
    \label{fig:unambiguous_compare_base}
}
\caption{
(a) The structure of the baseline network which is compared with the Text2Pickup network.
It is composed of a CNN module with three convolutional and max-pooling layers
and uses $e_s^c$, the sum of the word embedding vectors in the language command, as a language feature.
(b)
The result of comparison between the baseline network and the Text2Pickup network.
The result shows that the Text2Pickup network
is more robust than the baseline network when the commands are ambiguous.
}
\label{fig:correct_base}
\end{figure}

In this section, 
we validate the performance of the Text2Pickup network
by comparing with the baseline network shown in Figure~\ref{fig:baseline}.
The baseline network predicts the target block position $\mathbf{p}=(\frac{x}{n_i}, \frac{y}{n_i})$
without using an Hourglass network \cite{hourglass} and RNN \cite{lstm}.
As shown in Figure~\ref{fig:baseline},
the baseline network consists of the CNN module composed of three convolutional and max-pooling layers
and uses the sum of the word embedding vectors
$e_s^c = \sum_{k=1}^{n_c}e_k^c$
as a language feature.

We compare the results when 994 unambiguous language commands
and 265 ambiguous language commands in the test dataset
described in Section~\ref{sec:data}
are given as inputs to each network.
For the Text2Pickup network, 
the final prediction of the target block position is set to the
location where the value of the position heatmap is the highest.
Regarding this, the generated position heatmap is used
after being resized to the size of the input image.
For the baseline network,
the final prediction of the target block position
is obtained by multiplying the image size $n_i$ to the generated $\mathbf{p}$.

The experiment is defined as successful
when the distance between the predicted block position and the ground truth position
is less than 20 pixels, which is the half of the block size when the image size is 256.
Figure~\ref{fig:unambiguous_compare_base} shows the comparison result.
The accuracy is $66.10\%$ when an unambiguous (or certain) language command is given to the baseline network,
but if the input language command is ambiguous,
the accuracy is significantly lowered to $5.66\%$.
On the other hand, 
the accuracy of the Text2Pickup network
is $98.49\%$ when an unambiguous language command is given,
and $45.66\%$ when the input language command is vague.
This result shows that the Text2Pickup network
is more robust to the ambiguous language commands than the baseline network.
We claim that the Text2Pickup network which takes an advantage of
an Hourglass network and RNN
is superior to the neural network which finds the target block
based on the regression method.

\subsection{Comparison between the Interactive Text2Pickup and Text2Pickup Network} \label{sec:exp_compare_2}
In this section, we compare the accuracy of predicting the target block
before and after the interaction.
Here, we assume that only ambiguous language commands are given.
Regarding this, the performances of the Interactive Text2Pickup (IT2P) network 
and a single Text2Pickup network are compared.

For the IT2P network,
we implement a simulator that can answer a question from the network instead of real human users.
This simulator informs the color of the object when the network asks the color related questions,
and the position information of the object when the network asks the position related questions.
If the network asks questions related to the attributes of the desired block,
the simulator answers `Yes'.
If the network asks `This one?' and show the predicted target block,
the simulator responses `Yes' if the indicated block is the desired one.
If the indicated block is not the desired one,
the experiment is considered as a failure.

We supplement additional conditions to make the experiment more realistic and rigorous.
If the information provided in the language command (e.g., `Pick up the red block')
is asked again from the network (e.g, `Red one?'),
the experiment is considered as unsuccessful.
In addition,
if the question is related to the object
that cannot be indicated by an ambiguous language command is asked,
the experiment is considered as unsuccessful.
For example in Figure~\ref{fig:dataset}, when the language command is `Pick up the purple block',
the experiment is considered as a failure if the network asks `Lower left one?'.
When the distance between the predicted block position and the ground truth position
is less than 20 pixels on the image size of 256,
the experiment is considered as successful.

\begin{figure}[t]
\centering
\includegraphics[width=\linewidth]{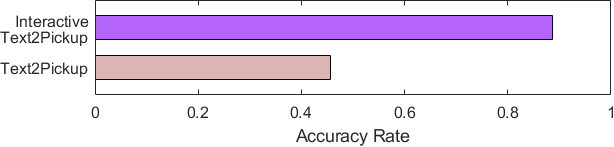}
\caption{
The result of comparison between the proposed Interactive Text2Pickup (IT2P) network
and a single Text2Pickup network
when ambiguous language commands are given.
The result shows that the proposed method can predict object location more accurately
by mitigating the ambiguity in the language command through the interaction.
}
\label{fig:interaction_compare}
\end{figure}

Figure~\ref{fig:interaction_compare} shows the comparison result before and after the interaction.
For a total of 265 ambiguous language commands in the test dataset described in Section~\ref{sec:data},
the accuracy of the proposed Interactive Text2Pickup (IT2P) network is $88.68\%$.
This shows that the proposed network which incorporates the human information gathered from the interaction
outperforms a single Text2Pickup network whose accuracy is $45.66\%$.
This high accuracy can be attributed to our simulator which provides the right information,
but this result obtained based the rigorous experimental conditions
shows that the proposed network can interact with humans effectively
by asking a question appropriate to the given situation.

\begin{figure*}[t]
\centering
\includegraphics[width=\linewidth]{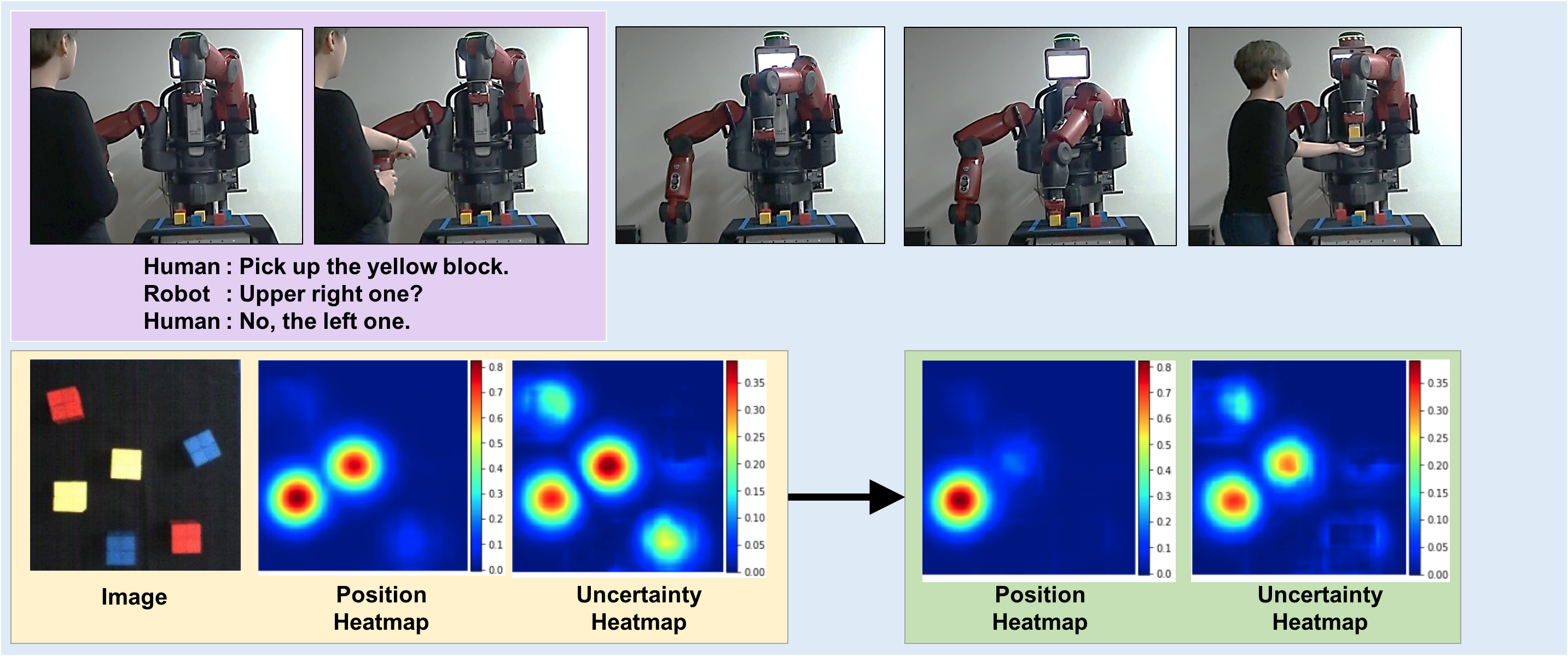}
\caption{
Experimental results of the
Interactive Text2Pickup (IT2P) network using a Baxter robot.
The subtitles in the purple rectangle show how the robot interacted with the human.
Heatmaps in the yellow rectangle were generated before the interaction,
and heatmaps in the green rectangle were generated after the interaction.
By obtaining more information from a human user, 
the robot succeeds in picking up the object that the human user requested.
}
\label{fig:exp_real}
\end{figure*}

\subsection{Experiments Using a Baxter Robot} \label{sec:exp_real}
Figure~\ref{fig:exp_real} shows the result of applying the proposed Interactive Text2Picktup (IT2P) network
to a Baxter robot.
The subtitles in the purple rectangle show how the robot interacts with the human user.
When a human user gives an ambiguous language command (``Pick up the yellow block'') to the robot, 
the heatmaps in the left yellow rectangle are generated.
The generated position heatmap points to the yellow blocks on the lower left and the upper right.
A high value of uncertainty is obtained around the red block,
because the trained network
has a difficulty in distinguishing between yellow and red,
due to the similarity in the RGB space.

By asking a question (``Upper right one?''),
the robot obtains additional information that the requested object is on the left,
and the generated heatmaps after the interaction are shown in the green rectangle in Figure~\ref{fig:exp_real}.
After the interaction,
the position heatmap indicates the yellow block on the left, 
and the value of the uncertainty heatmap is reduced overall.
Based on the interaction, the robot succeeds in picking up the requested object.

\section{Conclusion} \label{sec:conclud}
In this paper, we have proposed the Interactive Text2Pickup (IT2P) network
for picking up the requested object when a human language command is given.
The IT2P network interacts with a human user when an ambiguous language 
command is provided, in order to resolve the ambiguity.
By understanding the given language command,
the proposed network can successfully
predict the position of the desired object
and the uncertainty associated with the predicted target position.
In order to mitigate the ambiguity in the language command, 
the network generates a suitable question to ask the human user.
We have shown that the proposed IT2P network
can efficiently interact with humans
by asking a question appropriate to the given situation.
The proposed network it applied to a Baxter robot
and the collaboration between a real robot and a human user has been conducted.
We believe that the proposed method,
which can efficiently interact with humans by asking questions based on the estimation and the uncertainty, 
will enable more natural collaboration between a human and a robot.

\bibliographystyle{IEEEtran}
\bibliography{InteractivePickup}

\end{document}